\documentclass[12pt]{article}

\usepackage[margin=1in]{geometry}
\usepackage{amsfonts}       % blackboard math symbols
\usepackage{booktabs}       % professional-quality tables
\usepackage[utf8]{inputenc} % allow utf-8 input
\usepackage[T1]{fontenc}
\usepackage{macros}
\usepackage{microtype}      % microtypography
\usepackage{url}            % simple URL typesetting
\usepackage{hyperref}       % hyperlinks
\usepackage{nicefrac}       % compact symbols for 1/2, etc.
\usepackage{xcolor}         % colors

\usepackage{libertine}
\usepackage{libertinust1math}

\usepackage[round,authoryear]{natbib}

\hypersetup{
           breaklinks=true,   % splits links across lines
           colorlinks=true,   % displays links as colored text instead of blocks
}

\title{\LARGE \bfseries PAC Verification of \\ Statistical Algorithms}

\author{%
	{\normalsize Saachi Mutreja} \\
	{\small \textit{Computer Science Division}} \\
	{\small \textit{University of California, Berkeley}} \\
	{\small$\mathtt{saachi@berkeley.edu}$}
	\and
	{\normalsize Jonathan Shafer} \\
	{\small \textit{Computer Science Division}} \\
	{\small \textit{University of California, Berkeley}} \\
	{\small$\mathtt{shaferjo@berkeley.edu}$}
	\vspace*{0.5em}
}
\date{June 2023}

\begin{document}

\maketitle

\thispagestyle{empty}

\begin{abstract}
  \cite{GoldwasserRSY21} recently proposed the setting of PAC verification, where a hypothesis (machine learning model) that purportedly satisfies the agnostic PAC learning objective is verified using an interactive proof. In this paper we develop this notion further in a number of ways. First, we prove a lower bound of $\OmegaOf{\sqrt{d}/\varepsilon^2}$ i.i.d.\ samples for PAC verification of hypothesis classes of VC dimension $d$. Second, we present a protocol for PAC verification of unions of intervals over $\bbR$ that improves upon their proposed protocol for that task, and matches our lower bound's dependence on $d$. Third, we introduce a natural generalization of their definition to verification of general statistical algorithms, which is applicable to a wider variety of settings beyond agnostic PAC learning. Showcasing our proposed definition, our final result is a protocol for the verification of statistical query algorithms that satisfy a combinatorial constraint on their queries.
  
  %\todo{Showcasing our proposed definition, our final result is a protocol for the verification of top-down decision tree learning algorithms. To the best of our knowledge, this is the first result showing that a learning algorithm widely used in practice can be verified using an interactive proof system.}
\end{abstract}

\pagebreak

\thispagestyle{empty}

\tableofcontents

\vfill

% %%%%%%%%%%%%%%%%%%%%%%%%%%%%%%%%%%%%%%%%%%%%%%%%%%%%%%%%%%%%%%%%%%%%%%%%%%%%%%%%%%%%%%%%%%%%%%%%%%%%
% % TODO:
% %%%%%%%%%%%%%%%%%%%%%%%%%%%%%%%%%%%%%%%%%%%%%%%%%%%%%%%%%%%%%%%%%%%%%%%%%%%%%%%%%%%%%%%%%%%%%%%%%%%%
% \pagebreak
% \clearpage
% \thispagestyle{empty}
% \todo{
% 	\begin{itemize}
% 	\item{Related works: lower bound Thm 2 in Chiessa and Gur relates to our lower bound.}
% 	\item{Review \#422A: In Section 4, remind the reader how $m_V$ and $m_P$ are defined}
% 	\item{More comments in SODA reviews (/home/jonathan/Documents/PhD/My Papers/2023 SODA - PAC Verification of Statistical Algorithms/Review/).}
% 	\item{
% 		Bad math mode spacing in italics: Consider replacing all \$...\$ with \textbackslash(...\textbackslash) and using mathtools \texttt{mathic=true}, per \href{https://tex.stackexchange.com/questions/3490/bad-spacing-of-math-letters-within-italic-text}{this} answer.
% 	}
% 	\item{
% 		Add remark: counting SQ vs general SQ
% 	}
% 	\item{
% 		Add citation for ``the standard VC lower bound'' (\cite{VapChe68}??).
% 	}
% 	\end{itemize}
% }
% %%%%%%%%%%%%%%%%%%%%%%%%%%%%%%%%%%%%%%%%%%%%%%%%%%%%%%%%%%%%%%%%%%%%%%%%%%%%%%%%%%%%%%%%%%%%%%%%%%%%

\pagebreak

\clearpage
\pagenumbering{arabic} 

\section{Introduction}

Comparing what can be computed in a given model of computation versus what can be verified in that model is a recurring theme throughout the fields of computability and computational complexity. The most notorious example is of course the $\mathsf{P}$ vs.\ $\mathsf{NP}$ problem, which asks whether the set of decision problems that can be solved in polynomial time equals the set of decision problems whose solution can be verified in polynomial time given a suitable proof string. But the same question has been studied for many other settings and models of computation as well, with prominent examples including $\mathsf{L}$ vs.\ $\mathsf{NL}$ (for logspace computation), $\mathsf{P}$ vs.\ $\mathsf{IP} = \mathsf{PSPACE}$ (polytime computation, with an interactive proof) and $\mathsf{MIP}^* = \mathsf{RE}$ (ditto, with multiple quantum provers). The existence of a gap between computing and verifying is sometimes interpreted as capturing the notion of \emph{creativity}, in the sense that finding a solution to a problem might require discovery or inventiveness, while verifying a formal proof for the same is merely rote work.

While this theme has deep roots in the literature and an appealing interpretation, its parallels for \emph{learning} have only recently been explored for the first time. In the context of PAC\footnote{Probably Approximately Correct (PAC) is the standard theoretical model for supervised learning, introduced by \cite{VapChe68} and \cite{Valiant84}. Agnostic PAC learning is a generalization to the non-realizable case, introduced by \cite{Haussler92}. See also \cite{0033642}.} learning, \cite*{GoldwasserRSY21} introduced the setting of \emph{PAC verification}, in which an untrusted prover attempts to convince a verifier that a certain classifier has nearly-optimal loss with respect to a fixed unknown distribution from which the verfier can take random samples. Specifically, they work in the agnostic PAC setting, where the objective is to find a hypothesis $h$ that has nearly-optimal loss in the sense 
\begin{equation}\label{eq:pac-learning}
	\LossIndicator{\cD}{h} \leq \inf_{h' \in \cH} \LossIndicator{\cD}{h'} + \varepsilon,
\end{equation}
where $\LossIndicatorName{\cD}$ denotes $0$-$1$ population loss and $\cH$ is some fixed and known hypothesis class (formal definitions appear in \cref{section:preliminaries,section:verification-of-statistical-algorithm} below). 

Seeing as computational gaps are already well-studied, the main novelty in this setting concerns sample complexity gaps. They show that for some hypothesis classes (but not for others) the number of i.i.d.\ samples necessary to find a hypothesis with nearly-optimal loss is strictly greater than the number of i.i.d.\ samples necessary for verifying, with the help of an untrusted prover, that a proposed hypothesis has nearly-optimal loss.

Beyond the (substantial) theoretical motivation, this setting could have meaningful (and timely) real-world applications. First, if a sample complexity gap exists then ``verifiable data collection + ML as a service'' becomes a viable  business model. The provider would collect suitable training data from the desired population distribution, execute a chosen ML algorithm, and subsequently prove to the client that the end result is good with respect to the population distribution. The client would only need a small amount of independent data from the population distribution to determine the veracity of the claim. Beyond this, \cite{GoldwasserRSY21} envision a variety of other applications, such as more efficient schemes for replicating scientific results in the empirical sciences.

\subsection{Our Contributions}

PAC verification is novel territory, and very little is currently known. The current paper aims to make some modest steps towards charting this landscape. We focus on studying sample complexity gaps between learning and verifying specifically in terms of the dependence on the VC (Vapnik--Chervonenkis) dimension. 
%The VC dimension is one of three parameters that determine the sample complexity in classic PAC learning theory (the other two being the precision parameter $\varepsilon$ and confidence $\delta$), and we find it useful to focus our attention on this particular aspect.
We start with showing a lower bound for the sample complexity gap. Prior to our work, one could imagine that some classes would give rise to very large gaps, e.g., $\BigO{\log(d)}$ i.i.d.\ samples for verifying vs.\ the $\ThetaOf{d}$ samples that are known to be necessary and sufficient for learning, where $d = \VC{\cH}$. Our first result shows that the gap can be at most quadratic. Namely, for any hypothesis class, PAC verification requires that the verifier use at least $\OmegaOf{\sqrt{d}}$ i.i.d.\ random samples.

Second, we show that our lower bound's dependence on the VC dimension is tight in some cases, by improving upon a result of \cite{GoldwasserRSY21} to obtain a PAC verifier for the class of unions of intervals on $\bbR$ that uses $\BigO{\sqrt{d}}$ i.i.d.\ random samples. The previous result was an upper bound for a weaker notion of verification, that guarantees only that $\LossIndicator{\cD}{h} \leq 2\cdot \Opt + \varepsilon$, where $\Opt = \inf_{h' \in \cH} \LossIndicator{\cD}{h'}$ (instead of $\Opt + \varepsilon$ as in \cref{eq:pac-learning}). Their result applied only to a specific restriction of the class of unions of intervals, while our technique works for the restricted and for the unrestricted versions of the class.

Third, we take a step towards making the notion of PAC verification more applicable in practical settings. Many ML and data science algorithms that people use in practice, and might like to delegate to an untrusted service, do not obtain (or at least do not provably obtain) the objective of agnostic PAC learning as in \cref{eq:pac-learning}. Instead, they obtain some quantity of loss which is typically good enough in practice. With this reality in mind, we introduce a generalization of PAC verification that guarantees that the outcome is competitive with a specific algorithm. Namely, the verifier guarantees that with high probability, the hypothesis $h$ satisfies $\LossIndicator{\cD}{h} \leq \EE{\LossIndicator{\cD}{h_A}} + \varepsilon$, where $h_A$ is the (possibly randomized) output of the algorithm (see \cref{definition:pac-verification-of-algorithm}).

Fourth, we study PAC verification of statistical query algorithms. For a batch $\bq$ of statistical queries, we define a notion of \emph{partition size}, denoted $\partitionSize{\bq}$, which is the number of atoms in the $\sigma$-algebra generated by $\bq$. We show that whenever this quantity is sufficiently small, there is a sample complexity gap between execution and verification of the statistical query algorithm. 

Lastly, we show that there exists a sample complexity gap for a natural example we present, of optimizing a portfolio with advice. Both our lower bound and our upper bound apply to this example.

%\todo{Finally, we study PAC verification for top-down decision tree learning algorithms (such as ID3, CART and C4.5). This is an important example of a class of algorithms that are commonly used in practice and that do not obtain the agnostic PAC objective (see \cref{example:top-down-not-pac}). We show a verification protocol in which the verifier requires quadratically fewer i.i.d.\ samples than are necessary for executing the algorithm (see \todo{} \cref{theorem:admissability-verification,theorem:dp-decision-tree-verification}).}

\subsection{Related Works}

The study of interactive proofs for properties of distributions was initiated by \cite*{ChiesaG18}. They showed general bounds in terms of the support size. However, they did not consider tighter bounds that depend on combinatorial characterizations of the distribution testing property of interest (e.g., bounds that depend on the VC dimension).

The study of PAC verification of a hypothesis class was introduced by \cite*{GoldwasserRSY21}, who considered interactive proofs for properties of distributions in the specific context of machine learning. In particular, they also considered the relationship between the VC dimension of the class and the sample complexity of verification. They showed a lower bound that is incomparable with our lower bound, and they showed an upper bound for unions of intervals which is weaker than our upper bound. Our definition of PAC verification of an algorithm is closely modeled on their definition.

Recently, there have been a number of works on the general theme of distribution testing and interactive proofs for properties of distributions in the context of machine learning. These include \cite*{CanettiK21}, \cite*{abs-2108-12099}, \cite*{abs-2204-07196} and \cite*{HermanR22}, among others. \cite*{DBLP:journals/corr/abs-2306-04843} studied PAC verification with a quantum prover. \cite*{DBLP:journals/cacm/SeshiaSS22} survey the use of formal methods for verification of AI systems.

\subsection{Preliminaries}\label{section:preliminaries}

\begin{notation}
	$\bbN = \{1,2,3,\dots\}$, i.e., $0 \notin \bbN$. For any $n \in \bbN$, we denote $[n] = \{1,2,3,\dots,n\}$.
\end{notation}

\begin{notation}
	For a set $\Omega$, we write $\distribution{\Omega}$ to denote the set of all probability measures defined on the measurable space $(\Omega,\cF)$, where $\cF$ is some fixed $\sigma$-algebra that is implicitly understood. 
\end{notation}

\begin{definition}
	Let $\cP,\cQ$ be probability measures defined on a measurable space $(\Omega,\cF)$. The \ul{total variation distance} between $\cP$ and $\cQ$ is $\TV{\cP,\cQ} = \sup_{A \in \cF} |\cP(A)-\cQ(A)|$.
\end{definition}

%\begin{definition}
%	Let $\cX$ and $\cY$ be nonempty sets. A \ul{randomized function} $f: ~ \cX \rightarrow \cY$ is a tuple $(f^*,\Omega,\cR)$ such that $\Omega$ is a set, $f^*$ is a regular (deterministic) function $f^*: ~ \cX \times \Omega \rightarrow \cY$, and $\cR \in \distribution{\Omega}$ is a distribution. When $f$ is invoked with an input $x \in \cX$, a random value $r \sim \cR$ is sampled (independent from all previous samples from $\cR$), and the value $y = f^*(x,r) \in \cY$ is returned as output. In particular, repeated invocations of a randomized function with the same input yield i.i.d.\ outputs.
%\end{definition}

\subsubsection*{PAC Learning}

\begin{definition}\label{definition:VC}
	Let $\cX$ be a set, and let $\cH \subseteq \{0,1\}^\cX$ be a set of functions. Let $k \in \bbN$, $X = \{x_1,x_2,\dots,x_k\} \subseteq \cX$. We say that \ul{$\cH$ shatters $X$} if for any $y_1,y_2,\dots,y_k \in \{0,1\}$ there exists $h \in \cH$ such that  $h(x_i) = y_i$ for all $i \in [k]$. The \ul{Vapnik--Chervonenkis (VC) dimension of $\cH$}, denoted $\VC{\cH}$, is the largest $d \in \bbN$ for which there exist a set $X \subseteq \cX$ of cardinality $d$ that is shattered by $\cH$. If $\cH$ shatters sets of cardinality arbitrarily large, we say that $\VC{\cH} = \infty$.

\end{definition}

Throughout most of this paper we use loss functions of the type common in PAC learning, where the loss of a hypothesis with respect to a distribution is defined as the expected loss of that hypothesis on a randomly drawn sample form the distribution, as follows.

\begin{definition}\label{definition:weighted-loss}
	Let $\Omega$ and $\cH$ be sets. A \ul{loss function} is a function $\LossNameEmpty: ~ \Omega \times \cH \rightarrow [0,1]$. Let $h \in \cH$, and let $S = (z_1,\dots,z_m) \in \Omega^m$ be a vector. The \ul{empirical loss of $h$ with respect to $S$} is $\Loss{S}{h} = \frac{1}{m} \sum_{i \in [m]} \LossNameEmpty(z_i, h)$. For any distribution $\cD \in \distribution{\Omega}$, the \ul{loss of $h$ with respect to $\cD$} is $\Loss{\cD}{h} = \EEE{Z \sim \cD}{\LossNameEmpty(Z, h)}$. The \ul{loss of $\cH$ with respect to $\cD$} is $\Loss{\cD}{\cH} = \inf_{h' \in \cH} \Loss{\cD}{h'}$.
	
	The \ul{$0$-$1$ loss}, denoted $\LossIndicatorNameEmpty$, is the special case in which $\cX$ is a set, $\Omega = \cX \times \{0,1\}$, $\cH \subseteq \{0,1\}^\cX$, and $\LossNameEmpty((x,y),h) = \1(h(x) \neq y)$.
%
%
%	Let $\cX$ be a set, let $\cH \subseteq \{0,1\}^\cX$ be a class of hypotheses, let $\cD \in \distribution{\cX \times \{0,1}$, and let $h : ~ \cX \rightarrow \{0,1\}$ be a function. The \ul{$0$-$1$ loss of $h$ with respect to $\cD$} is $\Loss{\cD}{h} = \PPP{(x,y) \sim \cD}{h(x) \neq y}$.
%The \ul{$0$-$1$ loss of $\cH$ with respect to $\cD$} is $\Loss{\cD}{\cH} = \inf_{h' \in \cH}\Loss{\cD}{h'}$.
\end{definition}

However, in \cref{definition:pac-verification-of-algorithm} below we also consider more general types of loss. 

\begin{definition}
	Let $\cX$ be a set, and let $\cH \subseteq \{0,1\}^\cX$ be a class of hypotheses. We say that $\cH$ is \ul{agnostically PAC learnable} if there exist an algorithm $A$ and a function $\mA:~[0,1]^2 \rightarrow \bbN$ such that for any $\varepsilon,\delta \in (0,1)$ and any distribution ${\cD \in \Delta(\cX \times \{0,1\})}$, if $A$ receives as input a tuple of $\mA(\varepsilon,\delta)$ i.i.d.\ samples from $\cD$, then $A$ outputs a function $h \in \cH$ satisfying
	\[
	\PP{\LossIndicator{\cD}{h} \leq \LossIndicator{\cD}{\cH} + \varepsilon} \geq 1-\delta.
	\]
	In words, this means that $h$ is probably (with confidence $1-\delta$) approximately correct (has loss at most $\varepsilon$ worse than optimal). The point-wise minimal such function $\mA$ is called the \ul{sample complexity} of $\cH$.
\end{definition}

\subsubsection*{PAC Verification of a Hypothesis Class}

\begin{definition}[PAC Verification of a Hypothesis Class; a special case of \cite{GoldwasserRSY21}, Definition 4]\label{def:pac-verification}
	Let $\cX$ be a set, let ${\mathbb{D} \subseteq \Delta(\cX\times \{0,1\})}$ be a set of distributions, and let $\cH \subseteq \{0,1\}^{\cX}$ be a class of hypotheses. 
	We say that $\cH$ is \ul{PAC verifiable with respect to $\mathbb{D}$ using random samples} if 
	there exist an interactive proof system consisting of a verifier $V$ and an honest prover $P$ such that for any $\varepsilon,\delta \in (0,1)$ there exist $\mV,\mP \in \bbN$ such that for any $\cD \in \mathbb{D}$, the following conditions are satisfied:
	
	\begin{itemize}
		\item{
			\textbf{Completeness.} Let the random variable
			\[
			\hV = [V(S_V, \varepsilon, \delta), P(S_P, \varepsilon, \delta)] \in \cH \cup \{\reject\}
			\]
			denote the output of $V$ after receiving input $(S_V, \varepsilon, \delta)$ and interacting with $P$, which received input $(S_P, \varepsilon, \delta)$. Then
			\[
			\PPP{S_V \sim \cD^{\mV}, S_P \sim \cD^{\mP}}{\hV \neq \reject ~ \land ~ \Big(\LossIndicator{\cD}{\hV} \leq \LossIndicator{\cD}{\cH} + \varepsilon\Big)} \geq 1-\delta.
			\]
		}
		\item{
			\textbf{Soundness.} For any (possibly malicious and computationally unbounded) prover $P'$ (which may depend on $\cD$, $\varepsilon$, and $\delta$), the verifier's output $\hV = [V(S_V, \varepsilon, \delta), P']$ satisfies 
			\[
			\PPP{S_V \sim \cD^{\mV}, S_P \sim \cD^{\mP}}{\hV = \reject ~ \lor ~ \Big(\LossIndicator{\cD}{\hV} \leq  \LossIndicator{\cD}{\cH} + \varepsilon\Big)} \geq 1-\delta.
			\]
		}
	\end{itemize}
	In both conditions, the probability is over the randomness of the samples $S_V$ and $S_P$, as well as the randomness of $V$, $P$ and $P'$.
\end{definition}

\section{Technical Overview}\label{section:technical-overview}

\subsection{Bounds for Verification of VC Classes}

Our first result is a lower bound for the number of i.i.d.\ random samples the verifier requires to successfully PAC verify a class.  

\begin{theorem}\label{theorem:lower-bound}
	There exist constants $C,c > 0$ as follows. Let $\varepsilon \in (0,1)$, $\delta = 1/3$, let $\cX$ be a set, and let $\cH \subseteq \{0,1\}^\cX$ be a hypothesis class with $\VC{\cH} = d \in \bbN$. Assume that $(V,P)$ is an interactive proof system that PAC verifies $\cH$ with parameters $(\varepsilon,\delta)$ with respect to the set of all distributions $\bbD = \distribution{\cX \times \{0,1\}}$, and the verifier $V$ uses $\mV = \mV(d, \varepsilon)$ i.i.d.\ labeled samples. Then $\mV(d, \varepsilon) \geq (C \cdot \sqrt{d} - c)/\varepsilon^2$.
\end{theorem}

\begin{proof}[Proof Idea]
	This is an application of Le Cam's `point vs.\ mixture' method \citep[see][]{yu1997assouad}, together with a reduction from distribution testing to PAC verification. 
	Consider distributions where the marginal over the domain is uniform on a fixed $\cH$-shattered set of size $d$.
	PAC verification requires distinguishing the case of truly random labels (where the loss of the class is $\sfrac{1}{2}$), from the case where the labels are $\varepsilon$-biased (and the loss of the class is $\sfrac{1}{2}-\varepsilon$). An $\OmegaOf{\sqrt{d}/\varepsilon^2}$ lower bound for distinguishing these two cases is due to \cite{DBLP:journals/tit/Paninski08}.
\end{proof}

Our second result shows that the lower bound's dependence on $d$ is tight for a specific class.

\begin{theorem}\label{theorem:verification-of-d-intervals}
	Let $d \in \bbN$, and let 
	\[
		\cH_d = \left\{\1_X: ~ X = \bigcup_{i \in [d]}[a_i,b_i] ~ \land ~ \left(\forall i \in [d]: ~ 0 \leq a_i \leq b_i \leq 1 \right)\right\} \subseteq \{0,1\}^{[0,1]}
	\]
	be the class of boolean-valued functions over the domain $[0,1]$ that are indicator functions for a union of $d$ intervals. There exists an interactive proof system that PAC verifies the class $\cH_d$ with respect to the set of all distributions over $[0,1]\times \{0,1\}$, such that the verifier uses $\mV = \BigO{\sqrt{d}\log(\sfrac{1}{\delta})\varepsilon^{-2.5}}$ random samples, the honest prover uses
	\[
		\mP = \BigO{(d^2\log(d/\varepsilon) + \log(1/\delta))\varepsilon^{-4}}
	\]
	random samples, and both the verifier and the honest prover run in time polynomial in their numbers of samples.
\end{theorem}

\begin{proof}[Proof Idea]
	A discretization of the population distribution is induced by partitioning the domain $[0,1]$ into $d/\varepsilon$ intervals, each of which has weight $\varepsilon/d$ according to the population distribution. In the discretized distribution, the probability mass from each interval is lumped together into a single arbitrary point in that interval. We show that to find an $\varepsilon$-sub-optimal union of intervals, it suffices to know this discretized distribution. The prover sends the (purported) discretized distribution to the verifier. The verifier uses a distribution identity tester to verify that the provided distribution is a correct discretization of the population distribution. This is possible using $\BigO{\sqrt{d}}$ samples, because the support of the discretized distribution is of size $\BigO{d}$.
\end{proof} 

\subsection{Verification of Statistical Algorithms}\label{section:verification-of-statistical-algorithm}

Many popular algorithms do not come with provable PAC-like guarantees, but tend to work well in practice. Such heuristics are common in machine learning, data science, optimization, operations research, finance, etc. People might like to delegate the task of collecting data and executing an algorithm on that data to an untrusted party.
To capture this notion, our next contribution is a new definition of PAC verification of an \emph{algorithm}.\footnote{This notion differs from delegation of computation, in that the data (the input to the algorithm) is collected by the untrusted prover.}
This generalizes the definition of PAC verification of a \emph{hypothesis class} \citep[\cref{def:pac-verification}, introduced by][]{GoldwasserRSY21}, which corresponds to the special case of PAC verifying an algorithm that is an agnostic PAC learner for the class.

\begin{definition}[PAC Verification of an Algorithm]\label{definition:pac-verification-of-algorithm}
	Let $\Omega$ be a set, let $\mathbb{D} \subseteq \distribution{\Omega}$ be a set of distributions, 
	let $\cH$ be a set (called the set of \emph{possible outputs}), and for each $\cD \in \bbD$ let $\cO_\cD$ be an oracle.
	Let $A$ be a (possibly randomized) algorithm that takes no inputs, has query access to $\cO_\cD$, and outputs a value $\hA = A^{\cO_\cD} \in \cH$.
	Let $\LossNameEmpty: ~ \mathbb{D} \times (\cH\cup \{\reject\}) \rightarrow [0,1]$ be an arbitrary function\footnote{Note that this is more general than in \cref{definition:weighted-loss}.}, let $\Loss{\cD}{\cdot}$ denote $\LossNameEmpty(\cD, \cdot)$, and let $\Loss{\cD}{A} = \EE{\Loss{\cD}{\hA}}$, where the expectation is over the randomness
	of $A$ and of the oracle $\cO_\cD$.
	We say that the \ul{algorithm $A$ with access to oracles $\{\cO_\cD\}_{\cD \in \bbD}$ is PAC verifiable with respect to $\mathbb{D}$ by a verification protocol that uses random samples} if 
	there exist an interactive proof system consisting of a verifier $V$ and an honest prover $P$ such that for any $\varepsilon,\delta \in (0,1)$ there exist $\mV,\mP \in \bbN$ such that for any $\cD \in \mathbb{D}$, the following conditions are satisfied:
	\begin{itemize}
		\item{
			\textbf{Completeness.} Let the random variable
			\[
			\hV = [V(S_V, \varepsilon, \delta), P(S_P, \varepsilon, \delta)] \in \cH \cup \{\reject\}
			\]
			denote the output of $V$ after receiving input $(S_V, \varepsilon, \delta)$ and interacting with $P$, which received input $(S_P, \varepsilon, \delta)$. Then
			\[
			\PPP{S_V \sim \cD^{\mV}, S_P \sim \cD^{\mP}}{\hV \neq \reject ~ \land ~ \Loss{\cD}{\hV} \leq \Loss{\cD}{A} + \varepsilon} \geq 1-\delta.
			\]
		}
		\item{
			\textbf{Soundness.} For any deterministic or randomized (possibly malicious and computationally unbounded) prover $P'$ (which may depend on $\cD$, $\varepsilon$, $\delta$ and $\{\cO_\cD\}_{\cD \in \bbD}$), the verifier's output $h = [V(S_V, \varepsilon, \delta), P']$ satisfies 
			\[
				\PPP{S_V \sim \cD^{\mV}}{\hV = \reject ~ \lor ~ \Loss{\cD}{\hV} \leq \Loss{\cD}{A} + \varepsilon} \geq 1-\delta.
			\]
		}
	\end{itemize}
	The probabilities are over the randomness of $V$, $P$ and $P'$ and of the samples $S_V$ and $S_P$.
\end{definition}

In other words, whereas the definition of \cite{GoldwasserRSY21} required that the interactive proof system guarantee that a hypothesis is competitive with respect to any hypothesis in $\cH$, our definition requires that it be competitive with respect to a specific algorithm.  

\begin{remark}
	% Two natural candidate definitions for PAC verification of an algorithm require that with high probability, if the verifier does not reject, then either (1) $\Loss{\cD}{h} \leq \Loss{\cD}{h_A} + \varepsilon$, or (2) $\Loss{\cD}{h} \leq \EE{\Loss{\cD}{\hA}} + \varepsilon$. Candidate (1) requires that with high probability the verifier's output be at most $\varepsilon$ worse than the output of executing algorithm $A$, while (2) requires that it be at most $\varepsilon$ worse than the expected loss of $A$.
	PAC verification of an algorithm $A$ requires that $\Loss{\cD}{\hV} \leq \Opt_A + \varepsilon$ with high probability. Two natural candidate definitions for $\Opt_A$ include (1) $\Opt_A = \Loss{\cD}{h_A}$, and (2) $\Opt_A = \EE{\Loss{\cD}{\hA}}$. Candidate (1) requires that with high probability the verifier's output be at most $\varepsilon$ worse than the output of executing algorithm $A$, while (2) requires that it be at most $\varepsilon$ worse than the expected loss of~$A$.

	The loss $\Loss{\cD}{h_A}$ is a random variable that depends, inter alia, on the random samples used by $A$ (more generally: on the randomness of the oracle used by $A$). A crucial aspect of PAC verification is that the verifier use less random samples than are necessary for executing $A$, and in particular it cannot access the random samples used by $A$. So the verifier cannot know what loss was obtained in any particular execution of $A$. Therefore, we reject candidate (1) and adopt candidate (2).
\end{remark}

As an application of this new definition, we show that some statistical query algorithms (see \cref{definition:SQ-orcale,definition:sq-algorithm}) can be PAC verified via a protocol in which the verifier uses less i.i.d.\ samples than would be required for simulating the statistical query oracle used by the algorithm. Specifically, for a batch $\bq$ of statistical queries, the \emph{partition size} $\partitionSize{\bq}$ is the number of atoms in the $\sigma$-algebra generated by $\bq$. If the algorithm uses only batches with small partition size then verification is cheap, as in the following theorem.

\begin{reptheorem}{theorem:sq-verification}[Informal version]
	Let $A$ be a statistical query algorithm that adaptively generates at most $b$ batches of queries with precision $\tau$ such that each batch $\bq$ satisfies $\partitionSize{\bq} \leq s$.	Then $A$ is PAC verifiable by an interactive proof system where the verifier uses
	\[
		\mV = \ThetaOf{\frac{\sqrt{\psUpperBound}\log(\batchUpperBound/\varepsilon\delta)}{\tau^2} + \frac{\log(1/\varepsilon\delta)}{\varepsilon^2}}
	\]
	i.i.d.\ samples.
\end{reptheorem}

\begin{proof}[Proof Idea]
	The verifier simulates algorithm $A$. Each time $A$ sends a batch of queries to be evaluated by the statistical query oracle, the verifier sends the queries to the prover, and the prover sends back a vector of purported evaluations. The verifier uses $\BigO{\sqrt{s}/\tau^2}$ i.i.d.\ random samples to execute a distribution identity tester (\cref{theorem:tolerant-identity-testing}) to verify that the prover's evaluations are correct up to the desired accuracy $\tau$.
\end{proof}

In particular, \cref{theorem:sq-verification} implies the following separation:

\begin{repcorollary}{corollary:sq-verification-gap}[Informal version]
	Let $d \in \bbN$ and let $A$ be a statistical query algorithm such that each batch of queries generated by $A$ corresponds precisely to a $\sigma$-algebra with $d$ atoms. Then simulating $A$ using random samples requires $\OmegaOf{d/\tau^2}$ random samples, but there exists a PAC verification protocol for $A$ where the verifier uses $\BigO{\sqrt{d}/\tau^2}$ random samples.
\end{repcorollary}

\subsection{Examples}

%The following is a minimal example of a verification task in our generalized setting of \cref{definition:pac-verification-of-algorithm}, which is not an instance of \cref{def:pac-verification}, i.e., not an instance of PAC verification as defined by \cite{GoldwasserRSY21}.\footnote{In particular, the distribution in this task has no labels.} 

\begin{example}[Optimizing a portfolio with advice]
	Consider a task in which an agent selects a subset $S$ consisting of $n$ items from the set $\Omega = [2n]$. Subsequently, an item $i \in \Omega$ is chosen at random according to a distribution $\cD \in \distribution{\Omega}$ that is unknown to the agent, and the agent experiences loss $L(i,S) = \1(i \notin S)$.
	
	To help make an optimal decision, the agent has access to an i.i.d.\ sample $Z = (z_1,\dots,z_m) \sim \cD^m$. Let $\cH = \binom{\Omega}{n}$ denote the collection of subsets of size $n$ that the agent could select. $\VC{\cH} = n$, and therefore estimating the expected loss $\Loss{\cD}{S}$ of each possible choice $S \in \cH$ up to precision $\varepsilon > 0$ requires $\mA = \OmegaOf{(n + \log(1/\delta))/\varepsilon^2}$ samples.
	
	By \cref{corollary:sq-verification-gap}, if the agent can receive advice from an untrusted prover, it can make an $\varepsilon$-optimal choice using $\mV = \BigO{\sqrt{n}\log(1/\delta\varepsilon)/\varepsilon^2}$ i.i.d.\ samples. Note that $\mV \ll \mA$ for large $n$. Furthermore, our expression for $\mV$ is tight in the sense that, by \cref{theorem:lower-bound}, $\OmegaOf{\sqrt{n}}$ samples are necessary for verifying the advice of an untrusted prover.~\qed
\end{example}

Note that the above example is an instance of verification in our generalized setting (\cref{definition:pac-verification-of-algorithm}), but it is technically not an instance of PAC verification as previously defined by \cite{GoldwasserRSY21}, e.g., because the distribution has no labels. More generally, \cref{definition:pac-verification-of-algorithm} includes verification of \emph{distribution learning}, as follows. 

\begin{example}[Verification of distribution learning]\label{example:verification-distribution-learning}
	Let $\Omega = [n]$. Consider a task in which an agent has access to an i.i.d.\ sample $Z = (z_1,\dots,z_m) \sim \cD^m$ from some distribution $\cD \in \distribution{\Omega}$ that is unknown to the agent. The agent selects a distribution $\hat{\cD} \in \distribution{\Omega}$, and experience loss $\Loss{\cD}{\hat{\cD}} = \TV{\hat{\cD}, \cD}$.

	It is well known that to achieve loss at most $\varepsilon$ with probability at least $1-\delta$, it is necessary and sufficient to take $\mA = \ThetaOf{(n + \log(1/\delta))/\varepsilon^2}$ samples \cite[Theorem 1]{CanonneDistLearning}. In contrast, if the agent has access to advice from an untrusted prover then $\mV = \BigO{\sqrt{n}\log(1/\delta)\varepsilon^{-2}}$ i.i.d.\ samples are sufficient. The honest prover simply sends the verifier a description of a distribution $\tilde{\cD} \in \distribution{\Omega}$ that has loss at most $\varepsilon/\sqrt{n}$. The verifier uses distribution testing (\cref{theorem:tolerant-identity-testing}) to decide whether $\Loss{\cD}{\tilde{\cD}} \leq \varepsilon/\sqrt{n}$ or $\Loss{\cD}{\tilde{\cD}} \geq \varepsilon$, and accepts if and only if the former case holds.~\qed
\end{example}

A large collection of concrete tasks that might be of interest and that fall within the setting of \cref{definition:pac-verification-of-algorithm} involve solving various problems on graphs given random samples that convey information about the graph, as follows.  

\begin{example}[Verification in graphs]\label{example:graph-tasks}
	Fix $n \in \bbN$. For any graph $G = (V,E)$ with $V = [n]$, let $\cD_G$ be the uniform distribution on $E$. The agent does not know $G$, but it knows $n$ and it has access to an i.i.d.\ sample $Z = (z_1,\dots,z_m) \sim \cD_G^m$. Consider some standard tasks, such as:
	\begin{itemize}
		\item{
			Maximum matching. The agent selects a subset $M \subseteq \binom{V}{2}$ and experiences loss
			\[
				\Loss{\cD_G}{M} = \min_{M' \in \cM} \frac{|M \Delta M'|}{n},
			\]
			where $\cM$ is the set of all matchings in $G$ of maximal size.
		}
		\item{
			Coloring. The agent selects a function $f: ~ V \rightarrow \bbN$ and experiences loss
			\[
				\Loss{\cD_G}{f} = \min_{f' \in \cF} \frac{\sum_{v \in V}\1\left(f(v) \neq f'(v)\right)}{n}
			\]
			where $\cF$ is the set of all valid colorings of $G$ that use a minimal number of colors.
		}
	\end{itemize}

	For these tasks, there is an easy lower bound of $m = \OmegaOf{n}$ on the number of samples the agent needs to guarantee loss at most $\varepsilon$ with probability at least $1-\delta$ for $\varepsilon = \delta = 0.1$. 
	%To see this for maximum matching, consider a graph that is a disjoint union of pairs of vertices, such that in half of these pairs there exists an edge between the vertices (but the agent doesn't know in advance which pairs have a vertex and which do not). Finding an approximately maximum matching requires seeing nearly all the vertices in the graph.
	To see this, consider the family of graphs that consist of a disjoint union of triplets (sets of three vertices), such that each triplet contains a single edge. Because the agent does not know in advance where the edge is in each triplet, finding an approximately maximum matching and an approximate $2$-coloring require seeing nearly all the edges in the graph.

	However, if we assume that $G$ has maximum degree bounded by a constant (as in the lower bound), then $\cD_G$ is a uniform distribution with support size $\BigO{n}$. Hence, given access to advice from an untrusted prover, the agent can solve these tasks using $\BigO{\sqrt{n}}$ samples using the verification procedure of \cref{example:verification-distribution-learning}. 
	
	To see that $\OmegaOf{\sqrt{n}}$ samples are necessary for verification with the help of a prover, consider a family of graphs consisting of a disjoint union of triplets as above, but where only half the triplets contain an edge. Distinguishing between this family and the previous family requires observing a collision (receiving a sample that contains the same edge twice), which requires $\OmegaOf{\sqrt{n}}$ samples by the `birthday paradox'.~\qed
\end{example}

So far, all our examples involved a quadratic gap between learning and verifying. However, larger gaps are possible if we make strong assumptions on the unknown distribution. One example of this, pointed out by \cite{GoldwasserRSY21}, is that the gap between learning and verifying for \emph{realizable} PAC learning is unbounded. Unbounded gaps can exist also for other tasks as well, as in the following example.

\begin{example}[Unbounded gap in a graph task]
	Let $n$, $G=(V,E)$, and $\cD_G$ be as in \cref{example:graph-tasks}. Consider the maximal matching tasks under the assumption that $E$ is a perfect matching. Again, there is an easy lower bound of $\OmegaOf{n}$ random samples to guarantee loss at most $\varepsilon$ with probability at least $1-\delta$ for $\varepsilon = \delta = 0.1$ without the help of a prover. To see this, consider a graph that is a disjoint union of sets of four vertices, where each such set contains two disjoint edges. Finding a perfect matching requires seeing an edge from each set.
	
	In contrast, $\mV=\BigO{\log(1/\delta)/\varepsilon}$ samples are sufficient given advice from an untrusted prover. The protocol is as follows. The prover sends $\tilde{E}$, which purportedly equals $E$. If $\tilde{E}$ is not a perfect matching then the verifier rejects. Then, the verifier takes $\mV$ samples from $\cD_G$, and accepts if and only if all the edges in the sample appear in $\tilde{E}$. For completeness, if $\tilde{E} = E$ then the verifier always accepts. For soundness, if $\left(|\tilde{E} \Delta E|\right)/n \geq \varepsilon$, then $\cD_G$ has weight $\OmegaOf{\varepsilon}$ on edges that are not in $\tilde{E}$, and so taking $\mV$ samples is sufficient to ensure that the verifier rejects with probability at least $1-\delta$.~\qed
\end{example}

For the maximum matching task, we have seen that under the assumption that $G$ has maximum degree bounded by a constant the sample complexity gap is quadratic, but that the gap is unbounded under the stronger assumption that $G$ is a perfect matching. We view this as a demonstration of the richness of this setting.

\section{A Lower Bound for PAC Verification of VC Classes}

\cref{theorem:lower-bound} is proved via a reduction from the following distribution testing lower bound.

\begin{theorem}[Reformulation of Theorem 4 in \citealp{DBLP:journals/tit/Paninski08}]
	\label{theorem:paninski}
	Let $d,t \in \bbN$ and let $\varepsilon \in (0,1)$. For every $\sigma \in \Sigma = \{\pm 1\}^d$, let $\cD_{\sigma,\varepsilon} \in \distribution{[2d]}$ be a distribution such that for all $i \in [d]$,
	\[
		\cD_{\sigma,\varepsilon}(2i-1) = \frac{1 + \sigma_i\cdot\varepsilon}{2d},	\quad \text{and} \quad \cD_{\sigma,\varepsilon}(2i) = \frac{1 - \sigma_i\cdot\varepsilon}{2d}.
	\]
	Let $\cD_{\Sigma,\varepsilon,t}$ be the distribution over $[2d]^t$ generated by selecting a vector $\sigma \in \Sigma$ uniformly at random, and then taking $t$ i.i.d.\ samples from $\cD_{\sigma,\varepsilon}$. Let $\cD_{U,t} = \uniform{[2d]}^t$ be the distribution over $[2d]^t$ generated by selecting $t$ i.i.d.\ uniform samples from $[2d]$. Then $\TV{\cD_{U,t}, \cD_{\Sigma,\varepsilon,t}} \leq \paninski{t,\varepsilon,d}$ for 
	\[
		\paninski{t,\varepsilon,d} =  \frac{1}{2}\cdot\left(\exp\left(\frac{t^2\varepsilon^4}{d}\right)-1\right)^{1/2}.
	\]
\end{theorem}

%\cref{theorem:paninski} implies that if $t = t(d,\varepsilon) = \LittleO{\sqrt{d}/\varepsilon^2}$ then $\TV{\cD_{U,t}, \cD_{\Sigma,\varepsilon,t}} = \LittleO{1}$. In particular, for fixed $\delta < \sfrac{1}{2}$, there exists a constant $C > 0$ such that any distribution tester that takes $t(d,\varepsilon)$ i.i.d.\ samples from an unknown distribution $\cD$ and distinguishes with probability at least $1-\delta$  between the case $\cD = \uniform{[2d]}$ and the case $\cD \in \{\cD_{\sigma,\varepsilon}:~ \sigma \in \Sigma\}$, must satisfy 
% \begin{equation}\label{eq:testing-lower-bound-for-D-sigma}
% 	t(d,\varepsilon) \geq C \cdot \sqrt{d}/\varepsilon^2.
% \end{equation}

The proof also uses the following well-known fact about maximal couplings (see e.g.\ Lemma 4.1.13 in \citealp{roch2023+modern}).

\begin{theorem}
	\label{theorem:maximal-coupling}
	Let $\Omega$ be a set, and let $p_X,p_Y \in \distribution{\Omega}$ be distributions. Then
	\[
		\TV{p_X,p_Y}
		=
		\inf \: 
		\Big\{
			\PP{X \neq Y}: 
			~
			(X,Y)
			\text{ is a joint distribution with marginals }
			X \sim p_X
			\text{ and }
			Y \sim p_Y
		\Big\}.
	\]
\end{theorem}

\begin{proof}[Proof of \cref{theorem:lower-bound}]
	Let $X = \{x_1,\dots,x_d\} \subseteq \cX$ be a set of size $d$ that is shattered by $\cH$ (such a set exists because $\VC{\cH} = d$). Let $\cD_U = \uniform{X \times \{0,1\}}$. 
	
	% For every $h \in \cH_X = \{0,1\}^X$, let $\cD_{h,4\varepsilon} \in \distribution{X \times \{0,1\}}$ be a distribution such that for all $i \in [d]$,
	% \[
	% 	\cD_{h,4\varepsilon}\big((x_i,0)\big) = \frac{1 + (-1)^{h(x_i)}\cdot4\varepsilon}{2d},	\quad \text{and} \quad \cD_{h,4\varepsilon}\big((x_i,1)\big) = \frac{1 - (-1)^{h(x_i)}\cdot4\varepsilon}{2d}.
	% \]
	For every $h \in \cH_X = \{0,1\}^X$, let $\cD_{h,4\varepsilon} \in \distribution{X \times \{0,1\}}$ be a distribution such that
	\[
		\forall \: (x,y) \in X \times \{0,1\}: ~ 
		\cD_{h,4\varepsilon}\big((x,y)\big) 
		% = \frac{1 + (-1)^{\1(h(x) \neq y)}\cdot4\varepsilon}{2d}.
		=
		\left\{
		\begin{array}{ll}
			(1 + 4\varepsilon)/2d & h(x) = y \\
			(1 - 4\varepsilon)/2d & h(x) \neq y
		\end{array}
		\right..
	\]
	% Let $T$ be a distribution tester that solves the promise problem `$\cD_U$ vs.\ $\{\cD_{\sigma,\varepsilon}: ~ h \in \cH_X\}$' with success probability at least $2/3$. Namely, $T$ takes $t$ i.i.d.\ samples from an unknown distribution $\cD$ and distinguishes with probability at least $2/3$ between the case $\cD = \cD_U$ and the case $\cD \in \{\cD_{\sigma,\varepsilon}: ~ h \in \cH_X\}$, where $\cD_{\sigma,\varepsilon}$ is as in \cref{theorem:paninski}. If neither of these two cases holds then no assumption is made on the behavior of $T$.
	Consider a (possibly randomized) testing algorithm $T$ that takes $t$ i.i.d.\ samples from an unknown distribution $\cD$ and decides correctly with probability at least $1-\beta$ whether $\cD = \cD_U$ or whether $\cD \in \{\cD_{h,4\varepsilon}: ~ h \in \cH_X\}$ (if $\cD$ is not one of these $|\cH_X|+1$ options then we make no assumptions regarding the behavior of $T$).
	
	%By \cref{theorem:paninski} and \cref{eq:testing-lower-bound-for-D-sigma}, any such tester $T$ must use at least
	Let $\cD_{U,t} = (\cD_{U})^t$ and let $\cD_{\cH_X,4\varepsilon,t}$ be the distribution generated by selecting $h \in \cH_X$ uniformly at random and then taking $t$ i.i.d.\ samples from $\cD_{h,4\varepsilon}$.
	By \cref{theorem:paninski}, $\TV{\cD_{U,t}, \cD_{\cH_X,4\varepsilon,t}} \leq \paninski{t,4\varepsilon,d}$.
	By \cref{theorem:maximal-coupling}, for every $\alpha > 0$ there exists a joint distribution $(S_U,S_\cH)$ such that $S_U \sim \cD_{U,t}$, $S_\cH \sim \cD_{\cH_X,4\varepsilon,t}$, and $\PP{S_U \neq S_\cH} \leq \paninski{t,4\varepsilon,d} + \alpha$.

	For any such $\alpha$ and $(S_U,S_\cH)$, no tester can distinguish with probability strictly greater than $1/2$ between $S_U$ and $S_\cH$ in the event where $S_U = S_\cH$. Hence, 
	\begin{align*}
		\beta
		\geq
		1/2 \cdot \PP{S_U = S_\cH}
		=
		1/2 \cdot(1 - \PP{S_U \neq S_\cH})
		\geq
		1/2 \cdot (1-\paninski{t,4\varepsilon,d} - \alpha).
	\end{align*}
	Taking $\alpha \to 0$ and rearranging yields
	\begin{equation}\label{eq:testing-lower-bound-for-Dh}
		t \geq \frac{\sqrt{d \cdot \ln(1 + (4\beta - 2)^2)}}{\varepsilon^2}.
	\end{equation}
	This establishes a lower bound on the sample complexity for the $\cD_U$ vs.\ $\{\cD_{h,4\varepsilon}: ~ h \in \cH_X\}$ distribution testing problem. 
	
	Next, we show a reduction from the distribution testing problem to PAC verification of $\cH$. Let $(V,P)$ be an interactive proof system that PAC verifies $\cH$ such that the verifier $V$ and honest prover $P$ use $\mV$ and $\mP$ i.i.d.\ samples from the unknown distribution respectively, and satisfy \cref{def:pac-verification} with parameters $\varepsilon$ and $\delta$, as in the statement of \cref{theorem:lower-bound}.
	Using $(V,P)$, we construct a tester $T$ for the $\cD_U$ vs.\ $\{\cD_{h,4\varepsilon}: ~ h \in \cH_X\}$ testing problem. Given sample access to an unknown distribution $\cD$ for the testing problem, $T$ operates as follows:
	\begin{enumerate}
		\item{
			Compute $\hV = [V(\cD),P(\cD_U)]$. Namely, simulate an execution of the PAC verification protocol as follows. Take a sample $S_V \sim \cD^{\mV}$ of $\mV$ i.i.d.\ samples from $\cD$, and take a sample $S_P \sim (\cD_U)^{\mP}$ of $\mP$ i.i.d.\ samples from $\cD_U$ (seeing as the specification of $\cD_U$ is completely known to $T$, $T$ can generate as many samples from $\cD_U$ as necessary using uniform random coins). Execute the PAC verification protocol such that $V$ receives input $S_V$, $P$ receives input $S_P$, and the output of the verifier at the end of the protocol is $\hV \in \cH \cup \{\reject\}$.   
		}
		\item{
			Take a sample $S_{\mathsf{test}} \sim \cD^\ell$ of $\ell = \left\lceil \ln(24)/2\varepsilon^2 \right\rceil < 3/\varepsilon^2$ i.i.d.\ samples from $\cD$.
		}
		\item{
			If $(\hV = \reject)  ~ \lor ~ (\hV \neq \reject ~ \land ~ \LossIndicator{S_{\mathsf{test}}}{\hV} \leq \sfrac{1}{2}-2\varepsilon)$ then output ``$\cD \in \{\cD_{h,4\varepsilon} : ~ h \in \cH_X\}$''. Otherwise, output ``$\cD = \cD_U$''. 
		}
	\end{enumerate}
	We argue that the tester $T$ defined in this manner solves the testing problem correctly with probability at least $\sfrac{7}{12}$. If $\cD = \cD_U$, then $\LossIndicator{\cD}{h} = \sfrac{1}{2}$ for any $h \in \cH$. In particular, if $\hV \neq \reject$ then $\LossIndicator{S_{\mathsf{test}}}{\hV} \geq \sfrac{1}{2} - \varepsilon$ with probability at least $\sfrac{11}{12}$ (by Hoeffding's inequality and the choice of $\ell$). Thus, if $\cD = \cD_U$ then $T$ outputs ``$\cD = \cD_U$'' with probability at least $\sfrac{11}{12}$. 
	
	Conversely, if $\cD = \cD_{h',4\varepsilon}$ for some $h' \in \cH_X$,
	then $\LossIndicator{\cD}{h} = \sfrac{1}{2}-4\varepsilon$ for $h \in \cH$ such that $h|_X = h'$. From the correctness of the PAC verification protocol, with probability at least $\sfrac{2}{3}$, either $\hV = \reject$, or $\LossIndicator{\cD}{\hV} \leq \sfrac{1}{2}-3\varepsilon$, and in that case with probability at least $\sfrac{11}{12}$, $\LossIndicator{S_{\mathsf{test}}}{h} \leq \sfrac{1}{2}-2\varepsilon$  (again by Hoeffding's inequality and choice of $\ell$). A union bound implies that if $\cD = \cD_{h',4\varepsilon}$ for some $h' \in \cH_X$ then $T$ outputs ``$\cD \in \{\cD_{h,4\varepsilon} : ~ h \in \cH_X\}$'' with probability at least $1 - \sfrac{1}{3} - \sfrac{1}{12} = \sfrac{7}{12}$.

	We conclude that $T$ correctly solves the $\cD_U$ vs.\ $\{\cD_{h,4\varepsilon}: ~ h \in \cH_X\}$ testing problem with probability at least $\sfrac{7}{12}$ using $t = \mV + \ell$ i.i.d.\ samples from the unknown distribution $\cD$. Plugging $\beta=5/12$ in \cref{eq:testing-lower-bound-for-Dh}, this implies that $\mV \geq (0.3 \cdot \sqrt{d}-3)/\varepsilon^2$, as desired.~\qedhere
\end{proof}

\begin{remark}
	A previous version of this paper (\citealp{older-arxiv-version}) presented a proof of an $\OmegaOf{\sqrt{d}}$ lower bound, without the dependence on $\varepsilon$. That proof uses a reduction to a simpler distribution testing lower bound based on the `birthday paradox' (instead of the Paninski bound), and it may be better suited for pedagogical expositions.
\end{remark}

\section{Verification of Unions of Intervals}

\begin{theorem}[{\citealt[][Theorem 1]{CanonneJKL22}\footnote{See also \cite{GoldreichR11} and the discussion following Theorem 5.4 in \cite{canonne2020survey}.}}]\label{theorem:tolerant-identity-testing}
	Let $\varepsilon,\delta \in (0,1)$, let $n \in \bbN$, and let $\cP, \tilde{\cP} \in \distribution{[n]}$ be distributions. There exists a tolerant distribution identity tester that, given a complete description of $\tilde{\cP}$ and $m = \BigO{\sqrt{n}\log(1/\delta)\varepsilon^{-2}}$ i.i.d.\ samples from $\cP$, satisfies the following:
	\begin{itemize}
		\item{
			{\bfseries Completeness.} If $\TV{\cP,\tilde{\cP}} \leq \varepsilon/\sqrt{n}$ then the tester accepts with probability at least $1-\delta$. 
		}
		\item{
			{\bfseries Soundness.} If $\TV{\cP,\tilde{\cP}} > \varepsilon$ then the tester rejects with probability at least $1-\delta$.
		}
	\end{itemize}
\end{theorem}

\begin{definition}\label{definition:epsilon-sample}
	Let $\varepsilon \in [0,1]$, let $\cX$ be a set and let $\cF \subseteq \{0,1\}^\cX$ be a set of functions. Let $\cD \in \distribution{\cX}$, and let $S \in \cX^m$ for some $m \in \bbN$. We say that \ul{$S$ is an $\varepsilon$-sample for $\cD$ with respect to~$\cF$} if
	\[
		\forall f \in \cF: ~ \left|\frac{\left|\left\{x \in S: ~ f(x) = 1\right\}\right|}{m} - \PPP{x \sim \cD}{f(x) = 1}\right| \leq \varepsilon.	
	\]
\end{definition}

\begin{theorem}[\citealp{VapChe68}\footnote{Cf.\ \cite{AlonS00}, Theorem 13.4.4.}]\label{theorem:eps-sample}
	Let $d \in \bbN$ and $\varepsilon,\delta \in (0,1)$. Let $\cX$ be a set and let $\cF \subseteq \{0,1\}^\cX$ be a set of functions with $\VC{\cF} = d$. Let $\cD \in \distribution{\cX}$, and let $S \sim \cD^m$, where
	\[
		m = \OmegaOf{\frac{d\log(\sfrac{d}{\varepsilon}) + \log(\sfrac{1}{\delta})}{\varepsilon^2}}.
	\] 
	Then with probability at least $1-\delta$, $S$ is an $\varepsilon$-sample for $\cD$ with respect to $\cF$.
\end{theorem}

\begin{proof}[Proof of Theorem~\ref{theorem:verification-of-d-intervals}]
	We show that Protocol~\ref{protocol:union-of-intervals} (in \cref{appendix:union-of-intervals}) satisfies the requirements of the theorem. For completeness, note that if the prover follows the protocol then $\tilde{P}_{j,0} + \tilde{P}_{j,1} = \sfrac{1}{k}$ for all $j$, so the verifier will never reject at the first `if' statement. Let $\cB = \{I_j \times \{y\}: ~ j \in [k] ~ \land ~ y \in \{0,1\}\}$, and let $\cF = \{\1_E: ~ E \in \sigma(\cB)\} \subseteq \{0,1\}^{[0,1]\times \{0,1\}}$. In words, $\cF$ is the set of indicator functions for events in the $\sigma$-algebra generated by $\cB$. $\VC{\cF} = 2k = \BigO{d/\varepsilon}$, so \cref{theorem:eps-sample} and the choice of $\mP$ imply that with probability at least $1-\delta/2$, $S_P$ is an $\varepsilon/(6\sqrt{2k})$-sample for $\cD$ with respect to $\cF$. By the definitions of total variation distance and of an $\varepsilon$-sample, this implies that $\PP{\TV{\cP,\tilde{\cP}} \leq \varepsilon/(6\sqrt{2k})} \geq 1-\delta/2$. From the completeness of the tester of \cref{theorem:tolerant-identity-testing} and a union bound we conclude that with probability at least $1-\delta$, the verifier does not reject. This establishes completeness.

	For soundness, consider two cases. 
	\begin{itemize}
		\item{
			The prover is too dishonest, such that $\TV{\cP,\tilde{\cP}} > \varepsilon/6$. Then by the soundness of the tester of \cref{theorem:tolerant-identity-testing}, the verifier rejects with probability at least $1-\delta/2$.
		} 
		\item{
			The prover is sufficiently honest, such that $\TV{\cP,\tilde{\cP}} \leq \varepsilon/6$. 
			%For any $h' \in \cH_d$, define $\LossIndicator{\cP}{h'} = \PPP{(j,y) \sim \cP}{h'(x_j^*) \neq y}$ where for each $j \in [k]$,  $x_j^*$ is some fixed arbitrary point in $I_j$ as in the protocol, and define $L_{\tilde{\cP}}$ similarly. 
			%Then if the verifier does not reject, it outputs a hypothesis $h \in \argmin_{h' \in \cH}\LossIndicator{\tilde{\cP}}{h'}$. 
			Then for any $h' \in \cH_d$,
			\begin{align}
				\left|\LossIndicator{\cD}{h'} - \LossIndicator{\tilde{\cP}}{h'}\right| &\leq \left|\LossIndicator{\cD}{h'} - \LossIndicator{\cP}{h'}\right| + \left|\LossIndicator{\cP}{h'} - \LossIndicator{\tilde{\cP}}{h'}\right|  \nonumber\\
				&\leq \left|\LossIndicator{\cD}{h'} - \LossIndicator{\cP}{h'}\right| + \varepsilon/6, \label{eq:LD-vs-LPtilde}
			\end{align}
			where the last inequality follows from $\TV{\cP,\tilde{\cP}} \leq \varepsilon/6$.
		
			Fix $h' \in \cH_d$. We argue that $\left|\LossIndicator{\cD}{h'} - \LossIndicator{\cP}{h'}\right| \leq \varepsilon/3$. 
			% $\cP$ is a discretization of $\cD$, where the probability mass of all points $(x,y)$ such that $x \in I_j$ is lumped together into the probability mass of the single point $(x_j^*,y)$. Each $h' \in \cH_d$ is a union of $d$ intervals. 
			Let $Q = \{x \in [0,1]: ~ h'(x) \neq h'(x^*) \}$, where for each $x \in [0,1]$, we define $x^* = x_j^*$ such that $x \in I_j$. Namely, $Q$ is the set of points for which applying the discretization procedure alters the output of $h'$. Then
			\begin{align}
				\left|\LossIndicator{\cD}{h'}-\LossIndicator{\cP}{h'}\right| &= \left|\PPP{(x,y) \sim \cD}{h'(x) \neq y}-\PPP{(x,y) \sim \cD}{h'(x^*) \neq y}\right| \nonumber\\
				&= \Big|\PPP{(x,y) \sim \cD}{h'(x) \neq y ~ \land ~ x \in Q} \nonumber\\
				& ~~~~~~~~ -\PPP{(x,y) \sim \cD}{h'(x^*) \neq y  ~ \land ~ x \in Q}\Big| \label{eq:restrict-to-Q}\\
				&\leq \cD(Q') \tagexplain{$Q' = Q \times \{0,1\}$}\nonumber\\
				&\leq \sum_{j \in [k]: ~ I_j \cap Q \neq \varnothing} \cD(I_j') \tagexplain{$I_j' = I_j \times \{0,1\}$} \nonumber\\
				&= \sum_{j \in [k]: ~ I_j \cap Q \neq \varnothing} \cP(I_j') \tagexplain{$\cD(I_j')=\cP(I_j')$}\nonumber\\
				&=\cP\left(\bigcup\left\{I_j': ~ I_j \cap Q \neq \varnothing\right\}\right) \nonumber\\
				&\leq \tilde{\cP}\left(\bigcup\left\{I_j': ~ I_j \cap Q \neq \varnothing\right\}\right) + \TV{\cP,\tilde{\cP}} \nonumber\\
				&\leq 2d/k + \TV{\cP,\tilde{\cP}} \label{eq:d-endpoints}\\
				&\leq 2d/k + \varepsilon/6 = \varepsilon/3, \label{eq:prover-honest}
			\end{align}
			where \cref{eq:restrict-to-Q} holds since the loss of $h'$ can differ between $\cD$ and $\cP$ only for points in $Q$; \cref{eq:d-endpoints} holds because $h'$ consists of $d$ intervals, which together have $2d$ endpoints, $I_j \cap Q \neq \varnothing$ only if $I_j$ contains one of these endpoints, and if the verifier did not reject then $\tilde{\cP}(I_j') = \sfrac{1}{k}$ for all $j$; finally \cref{eq:prover-honest} holds by the assumption (in the current case) that the prover is sufficiently honest.

			Combining \cref{eq:prover-honest} with \cref{eq:LD-vs-LPtilde} yields $\forall h' \in \cH_d: ~ \left|\LossIndicator{\cD}{h'} - \LossIndicator{\tilde{\cP}}{h'}\right| \leq \varepsilon/2$. This implies that a hypothesis $h$ that has minimum loss with respect to $\tilde{\cP}$ satisfies $\LossIndicator{\cD}{h} \leq \LossIndicator{\cD}{\cH} + \varepsilon$. 
		}
	\end{itemize}
	We conclude that regardless of the prover's behavior, with probability at least $1-\delta/2$ the verifier either rejects or outputs a hypothesis with excess loss at most $\varepsilon$, as desired.
\end{proof}

\begin{remark}
	The dependence of the tolerance parameter in \cref{theorem:tolerant-identity-testing} on the domain size is quadratic, namely the verifier accepts if $\TV{\cP,\tilde{\cP}} \leq \varepsilon/\sqrt{n}$. Notice that this affects the sample complexity of the honest prover but not of the verifier. For instance, if the tolerance was $\varepsilon/e^n$ instead of $\varepsilon/\sqrt{n}$, the verifier's sample complexity would remain unchanged. 
\end{remark}

\section{Discussion and Future Work}

In this paper, we have shown that $\OmegaOf{\sqrt{d}}$ samples are necessary for PAC verifying a class of VC dimension $d$, and furthermore, for some classes $\BigO{\sqrt{d}}$ samples are sufficient. In contrast, Lemma
4.1 in \cite{GoldwasserRSY21} states that there also exist VC classes where the sample complexity for verification
is $\tildeOmegaOf{d}$ under the assumption that the verifier is proper (outputs a hypothesis from the class), and we believe it is likely that there exist VC classes for which an $\tildeOmegaOf{d}$ lower bound holds for any verifier. 

Hence, it appears likely that the VC dimension does not characterize the sample complexity of PAC verification. In that case, finding an alternative combinatorial quantity that does characterize that sample complexity is an exciting open problem. 

A potentially easier problem is to devise upper bounds (PAC verification protocols) for specific classes of interest. For example, the main property of the thresholds class utilized in the proof of \cref{theorem:verification-of-d-intervals} is that it has low `surface area' or noise sensitivity \citep[cf.][]{BalcanBBY12}. Perhaps a similar proof technique could apply to other classes as well.

Additionally, we introduced a notion of PAC verification of an algorithm. We believe this is very natural definition, because many of the algorithms that people
might like to delegate in practice are not PAC learners, including unsupervised learning algorithms
(e.g., clustering and dimensionality reduction algorithms), and supervised algorithms that are not provably PAC learners (e.g., neural networks trained via SGD). Devising PAC verification protocols for specific algorithms of interest could be a rewarding endeavor.

\phantomsection

\addcontentsline{toc}{section}{Acknowledgments}

\subsubsection*{Acknowledgments}

{\small
	An initial version of the lower bound in \cref{theorem:lower-bound} resulted from a conversation with Lijie Chen and Guy Rothblum. 
	JS would like to thank 
	Shafi Goldwasser,
	Steve Hanneke,
	Bobby Kleinberg,
	Shay Moran,
	Ido Nachum,
	Guy Rothblum
	and
	Abhishek Shetty
	for helpful comments and suggestions. Part of this work was done while JS was visiting the Weizmann Institute of Science (hosted by Guy Rothblum), Cornell University (hosted by Bobby Kleinberg) and the Technion (hosted by Shay Moran). JS is grateful for their hospitality and support.

	This work was supported in part by DARPA (Defense Advanced Research Projects Agency) contract \#HR001120C0015, and the Simons Collaboration on the Theory of Algorithmic Fairness. Any opinions, findings and conclusions or recommendations expressed in this material are those of the author(s) and do not necessarily reflect the views of the Simons Foundation or DARPA.
}

% \cleardoublepage

\newpage
\phantomsection

\addcontentsline{toc}{section}{References}

% \setcitestyle{numbers}
\bibliographystyle{plainnat}

\bibliography{paper}

\newpage

\appendix

\phantomsection

\addcontentsline{toc}{section}{Appendices}

\section{Protocol for Unions of Intervals}\label{appendix:union-of-intervals}

\begin{protocol}[H]
	\begin{ShadedAlgorithmBox}
		{\bfseries Assumptions:} 
		\begin{itemize}	
			\item{$d,\sfrac{1}{\varepsilon} \in \bbN$ (this can always be achieved by making $\varepsilon$ smaller if necessary), $k = 12d/\varepsilon$.}
   			\item{$\mP = \BigO{(d^2\log(d/\varepsilon) + \log(1/\delta))\varepsilon^{-4}}$ is a multiple of $k$.}
   			\item{$\mV = \BigO{\sqrt{d}\log(1/\delta)\varepsilon^{-2.5}}$.}
			% \item{
			% 	$m = \BigO{(d + \log(1/\delta))/\varepsilon^{2}}$
			% }
			\item{
				$S_V \sim \cD^{\mV}$, $S_P \sim \cD^{\mP}$.
			}
			\item{
				$\cD \in \distribution{[0,1]\times \{0,1\}}$ is an unknown target distribution.
			}
		\end{itemize}

		\noindent\rule{\textwidth}{0.5pt}

		\vsp

		\textsc{Prover}$(S_P,\delta,\varepsilon)$:
		\begin{algorithmic}
			\State $I_1,I_2,\dots,I_k \gets$ a partition of $[0,1]$ into disjoint intervals such that $\cup_{i \in [k]} I_i = [0,1]$
			\State \hfil and $\forall j \in [k]: ~ |\{x_1^P,\dots,x_{\mP}^P\}\cap I_j| = \mP/k$. \hfil
			\vsp
			\For $j \in [k]$:
				\For $b \in \{0,1\}$:
					\State $\tilde{P}_{j,b} \gets |\{(x,y) \in S_P: ~ x \in I_j ~ \land ~ y = b\}| / \mP$
					\Comment {Counted as a multiset}
				\EndFor
			\EndFor
			\State {\bfseries send} $\left(I_1,\dots,I_k\right)$ and $\big(\tilde{P}_{j,y}\big)_{j \in [k], y \in \{0,1\}}$ to the verifier
		\end{algorithmic}

		\vsp

		\textsc{Verifier}$(S_V,\delta,\varepsilon)$:	
		\begin{algorithmic}
			% \State Execute a uniformity tester using sample $S_V$ to check that
			% \[
			% 	\PP{\TV{P_I,\uniform{[k]}} > \varepsilon} \leq \delta,
			% \]
			% \State where $P_I \in \distribution{[k]}$ is the distribution generated by sampling $(x,y) \sim \cD$ and then selecting $j \in [k]$ such that $x \in I_j$
			% \vsppp
			\State {\bfseries receive} $\left(I_1,\dots,I_k\right)$ and $\big(\tilde{P}_{j,y}\big)_{j \in [k], y \in \{0,1\}}$ from the prover
			\vsp
			\If $\exists j \in [k]$ s.t. $\tilde{P}_{j,0} + \tilde{P}_{j,1} \neq \sfrac{1}{k}$:
				\State {\bfseries output} $\reject$ and {\bfseries terminate} 
			\EndIf

			\vsp

			\State $x_1^*,\dots,x_k^* \gets$ arbitrary points such that $\forall j \in [k]: ~ x_j^* \in I_j$

			\vsp

			\State {\bfseries execute} the tester of \cref{theorem:tolerant-identity-testing} with parameters $\sfrac{\varepsilon}{6}$, $\sfrac{\delta}{2}$ where $\cP,\tilde{\cP} \in \Delta([0,1] \times \{0,1\})$ are as follows:
			\begin{itemize}
				\item[-]{
					$\cP$ is the distribution generated by sampling $(x,y) \sim \cD$ and then outputting $(x^*,y)$ where $x^* = x_j^*$ such that $x \in I_j$
				}
				\item[-]{
					$\tilde{\cP}$ is the distribution such that $\PP{(x_j^*,y)} = \tilde{P}_{j,y}$ for all $j \in [k],y \in \{0,1\}$
				}
			\end{itemize}
			\vsp
			\If {distribution identity tester rejects:}
				\State {\bfseries output} $\reject$ and {\bfseries terminate} 
			\EndIf
			\vsp
			%\State $S' \gets \tilde{\cP}^m$ 
			%\Comment{\parbox[t]{.35\linewidth}{Generate $m$ i.i.d.\ samples from $\tilde{\cP}$ using random coins}}
			%\State $S' \gets \{(x_j^*,y): ~ (j,y) \in S'\}$
			%\Comment{\parbox[t]{.35\linewidth}{Treated as a multiset; for each $j$, fix a single arbitrary $x_j^* \in I_j$}}
			\State $h \gets \argmin_{h' \in \cH_d} \LossIndicator{\tilde{\cP}}{h'}$
			%\Comment An ERM hypothesis
			\vsp
			\State {\bfseries output} $h$  
			
		\end{algorithmic}	
	\end{ShadedAlgorithmBox}
	\caption{Verification protocol for unions of $d$-intervals.}
	\label{protocol:union-of-intervals}
\end{protocol}

\section{Verification of Statistical Query Algorithms}\label{section:sq-algorithm-verification}

\subsection{Definitions}

\subsubsection{Statistical Query Algorithms}

\begin{definition}[\citealp{Kearns98}]\label{definition:SQ-orcale}
	Let $\Omega$ be a set, let $\cD \in \distribution{\Omega}$ be a distribution, and let $\tau \geq 0$. A \ul{statistical query} is an indicator function $q: ~ \Omega \rightarrow \{0,1\}$. An oracle $\cO$ is a \ul{statistical query oracle for $\cD$ with precision $\tau$}, denoted $\cO \in \SQ{\cD,\tau}$, if at each invocation, $\cO$ takes a statistical query $q$ as input and produces an arbitrary evaluation $\cO(q) \in [0,1]$ as output such that
	\begin{equation}\label{eq:sq-oracle-def}
		\big|\cO(q) - \EEE{X \sim \cD}{q(X)}\big| \leq \tau.
	\end{equation} 
	In particular, the oracle's evaluations may be adversarial and adaptive, as long as each of them satisfies \cref{eq:sq-oracle-def}.
\end{definition}

\begin{remark}
	The notion of PAC verification of an algorithm (\cref{definition:pac-verification-of-algorithm}) requires that the verifier's output be competitive with $\Loss{\cD}{A} = \EE{\Loss{\cD}{A^\cO}}$, the expected loss of algorithm $A$ when executed with access to oracle $\cO$. For this expectation to be defined, throughout this paper we only consider oracles whose behavior can be described by a probability measure. In particular, oracles may be adaptive and adversarial in a deterministic or randomized manner, but they cannot be arbitrary.
\end{remark}

\begin{definition}\label{definition:sq-algorithm}
	A \ul{statistical query algorithm} is a (possibly randomized) algorithm $A$ that takes no inputs and has access to a statistical query oracle $\cO$. At each time step $t = 1,2,3,\dots$:
	\begin{itemize}
		\item{
			$A$ chooses a finite batch $\bq_t = \left(q_t^1,\dots,q_t^{\numQueries_\timestep}\right)$ of statistical queries and sends it to the oracle $\cO$.
		}
		\item{
			$\cO$ sends a batch of evaluations $\bv_t = \left(v_t^1,\dots,v_t^{\numQueries_\timestep}\right) \in [0,1]^{\numQueries_\timestep}$ to $A$, such that $v_t^i = \cO(q_t^i)$ for all $i \in [\numQueries_\timestep]$.
		}
		\item{
			$A$ either produces an output and terminates, or continues to time step $t+1$.
		}
	\end{itemize}
	The resulting sequence $\br = (\bq_1,\bv_1,\bq_2,\bv_2,\dots)$ is called a \ul{transcript} of the execution.
\end{definition}

Note that for each $t$, the choice of $\bq_t$ is a deterministic function of $(\br_{< t},\rho)$, where 
\[
	\br_{< t} = (\bq_1,\bv_1,\bq_2,\bv_2,\dots,\bq_{t-1},\bv_{t-1}),
\]
and $\rho$ denotes the randomness of $A$. If $A$ terminates, its final output is a deterministic function of $(\br,\rho)$.

\subsubsection{The Partition Size}

\begin{definition}
	Let $\Omega$ be a set, and let $\cS \subseteq 2^\Omega$ be a collection of subsets. We say that $\cS$ is a \ul{$\sigma$-algebra for $\Omega$} if it satisfies the following properties:
	\begin{itemize}
		\item{$\Omega \in \cS$.}
		\item{$\forall S \subseteq \cS: ~ \Omega \setminus S \in \cS$.}
		\item{For any countable sequence $S_1,S_2,\ldots \in \cS: ~ \cup_{i=1}^\infty S_i \in \cS$.}
	\end{itemize}
\end{definition}

\begin{definition}
	Let $\Omega$ be a set.
	\begin{itemize}
		\item{
			Let $\cA \subseteq 2^\Omega$ be a collection of subsets. The \ul{$\sigma$-algebra generated by $\cA$ for $\Omega$}, denoted $\sigma(\cA)$, is the intersection of all $\sigma$-algebras for $\Omega$ that are supersets of $\cA$.
		}
		\item{
			Let $\cF \subseteq \{0,1\}^\Omega$ be a set of indicator functions. The \ul{$\sigma$-algebra generated by $\cF$ for $\Omega$} is $\sigma(\cF) = \sigma\left(\{A \subseteq \Omega: ~ \1_A \in \cF\}\right)$.
		}
	\end{itemize}
\end{definition}

\begin{definition}
	Let $\cS$ be a $\sigma$-algebra. The set of \ul{atoms of $\cS$} is 
	\[
		\atoms{\cS} = \left\{S \in \cS: ~ \left(\forall S' \in \cS\setminus\varnothing: ~ S' \not\subset S\right)\right\}.\footnote{$S' \not\subset S$ denotes that $S'$ is not a strict subset of $S$.}
	\]
\end{definition}

\begin{definition}\label{definition:PS}
	Let $\Omega$ be a set and let $\cF = \{f_1,f_2,\dots,f_k\} \subseteq \{0,1\}^\Omega$ be a finite set of indicator functions. The \ul{partition size of $\cF$} is $\partitionSize{\cF} = |\atoms{\sigma(\cF)}| \in \bbN$, i.e., the number of atoms in the $\sigma$-algebra generated by $\cF$ for $\Omega$.
\end{definition}

\subsection{Formal Statements}

\begin{theorem}[PAC Verification of an SQ Algorithm]\label{theorem:sq-verification}
	Let $\batchUpperBound, \psUpperBound \in \bbN$, let $\Omega$ be a set and $\cH$ be a discrete set. Let $A$ be a statistical query algorithm that adaptively and randomly generates some random number $\finalBatch$ of batches $\bq_1,\dots,\bq_{\finalBatch}$ of statistical queries $\Omega \rightarrow \{0,1\}$ such that with probability $1$, $\finalBatch \leq \batchUpperBound$ and $\partitionSize{\bq_t} \leq s$ for each $t \in [\finalBatch]$, and the algorithm outputs a random value $h \in \cH$. Let $\mathbb{D} \subseteq \distribution{\Omega}$ be a set of distributions, let $\tau > 0$, and let $\LossNameEmpty: ~ \Omega \times \cH \rightarrow [0,1]$ be a loss function.
	
	Then there exists a collection of oracles $\bbO = \{\cO_\cD\}_{\cD \in \bbD}$ where $\cO_\cD \in \SQ{\cD,\tau}$ for all $\cD \in \bbD$, such that algorithm $A$ with access to oracles $\bbO$ is PAC verifiable with respect to $\mathbb{D}$ by a verification protocol that uses random samples, where the verifier and honest prover respectively use
	\[
		\mV = \ThetaOf{\frac{\sqrt{\psUpperBound}\log(\batchUpperBound \numIterations/\delta)}{\tau^2} + \frac{\log(\numIterations/\delta)}{\varepsilon^2}},
	\]
	and
	\[
		\mP = \ThetaOf{\frac{\psUpperBound^3\log(\psUpperBound\batchUpperBound \numIterations/\delta\tau)}{\tau^2}}
	\]
	i.i.d.\ samples, with $\numIterations = \left\lceil 8\log(4/\delta)/\varepsilon \right\rceil$.
\end{theorem}

As a corollary, we obtain that for statistical query algorithms of a particular type, the sample complexity of PAC verification has a quadratically lower dependence on the VC dimension of the batches of statistical queries compared to simulating the algorithm using random samples.

\begin{corollary}\label{corollary:sq-verification-gap}
	Let $A$ be a statistical query algorithm as in \cref{theorem:sq-verification}, and let $d \in \bbN$. Assume that in each time step $t \in [\finalBatch]$, $\VC{\bq_t} = d$ and $|\bq_t| = 2^d$. Namely, $\bq_t$ is the set of indicator functions of a $\sigma$-algebra with $d$ atoms. Consider an implementation of $A$ that uses random samples to simulate the SQ oracle accessed by $A$, such that the implementation uses random samples only and does not use any oracles. Simulating an oracle $\cO \in \SQ{\cD,\tau}$ requires
	\[
		m = \OmegaOf{\frac{d + \log(1/\delta)}{\tau^2}}
	\]
	i.i.d.\ samples from $\cD$. In contrast, there exists a protocol that PAC verifies $A$ such that the verifier uses only
	\[
		\mV = \ThetaOf{\frac{\sqrt{d}\log(\batchUpperBound \numIterations/\delta)}{\tau^2} + \frac{\log(\numIterations/\delta)}{\varepsilon^2}}
	\]
	i.i.d.\ samples from $\cD$, with $\numIterations = \numIterationsExpression$.
\end{corollary}

The lower bound in the corollary is the standard VC lower bound.

\subsection{Proofs}

\begin{definition}
	Let $A$ be a statistical query algorithm, let $\bbD$ be a collection of distributions, and let $\varepsilon,\tau > 0$. We say that a collection of oracles $\bbO = \{\cO_\cD\}_{\cD \in \bbD}$ is \ul{$\varepsilon$-maximizing with respect to $A$ and $\bbD$} if for each $\cD \in \bbD$, $\cO_\cD \in \SQ{\cD,\tau}$ and $\EE{\Loss{\cD}{A^{\cO_\cD}}} \geq \sup_{\cO \in \SQ{\cD,\tau}}\EE{\Loss{\cD}{A^{\cO}}} - \varepsilon$.
\end{definition}

\begin{proof}[Proof of \cref{theorem:sq-verification}]
	Fix a collection of oracles $\bbO = \{\cO_\cD\}_{\cD \in \bbD}$ that is $\varepsilon/4$-maximizing with respect to $A$ and $\bbD$.
	We show that algorithm $A$ with access to the oracles $\bbO$ is PAC verified by \cref{protocol:sq-verification}.
	
	To establish completeness, notice that each batch $\ba_t$ of queries sent to the prover by \textsc{VerifierIteration} satisfies $\VC{\ba_t} = 1$, and there are at most $\batchUpperBound \cdot \numIterations$ such batches. Hence, by \cref{theorem:eps-sample} and a union bound, taking $\mP$ as in the statement is sufficient to guarantee that with probability at least $1-\scompletenessfrac{1}$, 
	%for all batches $\ba_t$, the query evaluations $\tilde{\bp}_t$ provided by the prover satisfy 
	\[
		\forall \text{ iteration } i \in [\numIterations] ~ \forall t \in [\finalBatch]: ~  \left\|\tilde{\bp}_t  - \bp_t\right\|_\infty \leq \frac{\tau}{\psUpperBound\sqrt{\psUpperBound}},
	\]
	where $\bp_t$ is the vector of correct evaluations, with components $p_t^j = \EEE{Z \sim \cD}{a_t^j(Z)}$. Hence, with probability at least $1-\scompletenessfrac{1}$,
	\begin{equation}\label{eq:p-tilde-close-to-p-for-learning}
		\forall \text{ iteration } i \in [\numIterations] ~ \forall t \in [\finalBatch]: ~  \left\|\tilde{\bp}_t  - \bp_t\right\|_1 \leq \frac{\tau}{\sqrt{\psUpperBound}}.
	\end{equation}
	By \cref{eq:p-tilde-close-to-p-for-learning}, \cref{theorem:tolerant-identity-testing}, and the choice of $\mV$, with probability at least $1-\scompletenessfrac{1}$, none of the executions of \textsc{IdentityTest} cause the verifier to reject.
	
	By a union bound, with probability at least $1-\scompletenessfrac{2}$, \cref{eq:p-tilde-close-to-p-for-learning} holds and the verifier does not reject. Then, by \cref{lemma:sq-verification-lemma}, 
	\begin{equation}\label{eq:h-is-good-with-pr-eps}
		\forall i \in [\numIterations]: ~ \PP{\Loss{\cD}{h_i} \leq \Loss{\cD}{A} + \frac{\varepsilon}{2}} \geq \frac{\varepsilon}{8}.
	\end{equation}
	By the choice of $\numIterations$,
	\begin{align}\label{eq:exists-good-h}
		\PP{\forall i \in [\numIterations]: ~ \Loss{\cD}{h_i} > \Loss{\cD}{A} + \frac{\varepsilon}{2}} &\leq \left(1-\frac{\varepsilon}{8}\right)^\numIterations \leq e^{-\varepsilon\numIterations/8} \leq \completenessfrac{1}.
	\end{align}
	
	By Hoeffding's inequality, a union bound, and the choice of $\mV$, 
	\begin{equation}\label{eq:uniform-convergence-for-h}
		\PP{\forall i \in [\numIterations]: ~ \left|\Loss{S_V'}{h_i} - \Loss{\cD}{h_i}\right| \leq \frac{\varepsilon}{2}} \geq 1-\completenessfrac{1}.
	\end{equation}
	Combining \cref{eq:p-tilde-close-to-p-for-learning,eq:exists-good-h,eq:uniform-convergence-for-h} via a union bound, we conclude that with probability $1-\scompletenessfrac{4}$, the verifier does not reject and it outputs $h \in \cH$ such that $\Loss{\cD}{h} \leq \Loss{\cD}{A} + \varepsilon$. This establishes completeness.
	
	To establish soundness, consider an interaction between the verifier of \cref{protocol:sq-verification} and any deterministic or randomized (possibly malicious and computationally unbounded) prover $P'$, and examine the following two events.
	
	\begin{itemize}
		\item{
			Event I: the evaluations provided by $P'$ satisfy 
			\begin{equation}\label{eq:p-tilde-close-to-p-for-verification}
				\forall \text{ iteration } i \in [\numIterations] ~ \forall t \in [\finalBatch]: ~  \left\|\tilde{\bp}_t  - \bp_t\right\|_1 \leq \tau.
			\end{equation}
			If the verifier does not reject then \cref{lemma:sq-verification-lemma} implies that \cref{eq:h-is-good-with-pr-eps} holds. As we saw in the proof for the completeness requirement, this implies that with probability at least $1-\scompletenessfrac{4}$, the verifier outputs $h \in \cH$ such that $\Loss{\cD}{h} \leq \Loss{\cD}{A} + \varepsilon$.
		
		}
		\item{
			Event II: there exists an iteration $i \in [\numIterations]$ containing a time step $t^* \in [\finalBatch]$ such that $\left\|\tilde{\bp}_{t^*}  - \bp_{t^*}\right\|_1 > \tau$. By \cref{theorem:tolerant-identity-testing} and the choice of $\mV$, with probability at least $1-\scompletenessfrac{1}$ the verifier rejects in time step $t^*$.
		}
	\end{itemize}
	We conclude that in both cases, 
	\[
		\PPP{S_V \sim \cD^{\mV}}{h = \reject ~ \lor ~ \Loss{\cD}{h} \leq \Loss{\cD}{A} + \varepsilon} \geq 1-\delta,
	\]
	and this establishes soundness.
\end{proof}

\begin{lemma}\label{lemma:sq-verification-lemma}
	In the context of \cref{theorem:sq-verification}, fix a distribution $\cD \in \bbD$ and let $\cO_\cD \in \SQ{\cD,\tau}$ be an oracle such that 
	\[
		\EE{\Loss{\cD}{A^{\cO_\cD}}} \geq \sup_{\cO \in \SQ{\cD,\tau}}\EE{\Loss{\cD}{A^{\cO}}} - \varepsilon/4.
	\] 
	Consider an execution of \textsc{VerifierIteration} (\cref{protocol:sq-verifier-iteration}).
	Let $G$ denote the event in which the verifier does not reject, and the query evaluations $\tilde{\bp}_t$ provided by the prover satisfy
	\begin{align}\label{eq:assumption-prover-honest-enough-for-learning}
		\forall t \in [\finalBatch]: ~ \left\|\tilde{\bp}_t  - \bp_t\right\|_1 \leq \tau,
	\end{align}
	where $\bp_t$ is the vector of correct evaluations $p_t^i = \EEE{Z \sim \cD}{a_t^i(Z)}$. Then the output $\hVI \in \cH$ returned by \textsc{VerifierIteration} satisfies
	\begin{equation}\label{eq:hVI-low-loss}
		\PP{\Loss{\cD}{\hVI} \leq \EE{\Loss{\cD}{A^{\cO_\cD}}} + \frac{\varepsilon}{2} ~ \Big\vert ~ G} \geq \frac{\varepsilon}{8}.
	\end{equation}
\end{lemma}

\begin{proof}
	Let $\cO_G$ denote the oracle with evaluations that are equal in distribution to the evaluations provided by the prover conditioned on event $G$ occurring. By the choice of $\cO_\cD$, 
	\begin{align*}
		\EE{\Loss{\cD}{\hVI} ~ | ~ G} = \EE{\Loss{\cD}{A^{\cO_G}}} \leq \EE{\Loss{\cD}{A^{\cO_\cD}}} + \varepsilon/4.
	\end{align*}
	By Markov's inequality, 
	\begin{align*}
		\PP{\Loss{\cD}{\hVI} > \EE{\Loss{\cD}{A^{\cO_\cD}}} + \varepsilon/2  ~ \big| ~ G} &\leq \PP{\Loss{\cD}{\hVI} > \EE{\Loss{\cD}{\hVI} ~ | ~ G} + \varepsilon/4  ~ \big| ~ G} \\
		&\leq \frac{\EE{\Loss{\cD}{\hVI} ~ | ~ G}}{\EE{\Loss{\cD}{\hVI} ~ | ~ G} + \varepsilon/4} \\
		&\leq \frac{1}{1+\varepsilon/4},
	\end{align*}
	since $\LossName{\cD}$ is at most $1$.
	Hence, the complement satisfies
	\begin{align*}
		\PP{\Loss{\cD}{\hVI} \leq \EE{\Loss{\cD}{A^{\cO_\cD}}} + \frac{\varepsilon}{2}  ~ \Big\vert ~ G} \leq \frac{\varepsilon/4}{1+\varepsilon/4} \leq \frac{\varepsilon}{8},
	\end{align*}
	as desired.
\end{proof}

\begin{protocol}[H]
	\begin{ShadedAlgorithmBox}
		{\bfseries Assumptions:} 
		\begin{itemize}
			\item{
				$\Omega$ is a set, $\cD \in \distribution{\Omega}$ is the population distribution.
			}
			\item{
				$A$ is a statistical query algorithm to be verified.
			}
			\item{
				$\tau \in (0,1)$ is the accuracy parameter for statistical queries used by $A$.
			}
			\item{
				$\batchUpperBound \in \bbN$ is an upper bound on the number of statistical query batches generated by $A$.
			}
			\item{
				$\varepsilon,\delta \in (0,1)$ are the desired accuracy and confidence parameters for the verification.
			}
			\item{
				$\numIterations = \numIterationsExpression$.
			}
			\item{
				$\mV = \ThetaOf{\sqrt{\psUpperBound}\log(\batchUpperBound \numIterations/\delta)\tau^{-2} + \log(\numIterations/\delta)\varepsilon^{-2}}$.
			}
			\item{
				$\mP = \ThetaOf{\psUpperBound^3\log(\psUpperBound\batchUpperBound \numIterations/\delta\tau)\tau^{-2}}$.
			}	
			\item{
				$S_{V},S_{V}' \sim \cD^{\mV}$, $S_P \sim \cD^{\mP}$ are independent sets of i.i.d.\ samples.
			}
			\item{
				$S_{V} = (\samplepoint{1}{V},\dots,\samplepoint{\mV}{V})$, $S_{V}' = (\samplepoint{1}{V'},\dots,\samplepoint{\mV}{V'})$, $S_P = (\samplepoint{1}{P},\dots,\samplepoint{\mP}{P})$.
			}
			%			\item{
			%				$\lambda = \tau \batchUpperBound/\ln(2)$.
			%			}	
			%			\item{
			%				$\varepsilon,\delta,\tau \geq 0$.
			%			}
			%%			\item{
			%%				$\tau > 0$ is the precision of the $\SQ{\cD,\tau}$ oracle of the algorithm to be verified.
			%%			}
			%			\item{
			%				$s = \partitionSize{A,\bbO}$
			%			}
			%			\item{
			%				$\mV = \ThetaOf{\sqrt{s}\log(1/\varepsilon\delta)\tau^{-4}}$
			%			}
			%			\item{
			%				\todo{} $\mV'\mP$.
			%			}
			%			\item{
			%				$S_V = \big((x_1^{\scriptscriptstyle{V}},y_1^{\scriptscriptstyle{V}}),\dots,(x_{\mV}^{\scriptscriptstyle{V}},y_{\mV}^{\scriptscriptstyle{V}})\big)\sim \cD^{\mV}$ are i.i.d.\ random samples.
			%			}
			%			\item{
			%				$S_V' \sim \cD^{\mV'}$ are i.i.d.\ random samples.
			%			}
			%			\item{
			%				$S_P = \big((x_1^{\scriptscriptstyle{P}},y_1^{\scriptscriptstyle{P}}),\dots,(x_{\mP}^{\scriptscriptstyle{P}},y_{\mP}^{\scriptscriptstyle{P}})\big)\sim \cD^{\mP}$ are i.i.d.\ random samples.
			%			}
		\end{itemize}
		
		\noindent\rule{\textwidth}{0.5pt}
		
		\vspace*{0.5em}
		
		\textsc{Verifier}$(S_V, S_V')$:
		\begin{algorithmic}
			\For $i \in [\numIterations]$: 
			\State $h_i \gets \textsc{VerifierIteration}(S_V)$
			\Comment{\cref{protocol:sq-verifier-iteration}}
			\EndFor
			\State $i^* \gets \argmin_{i \in [\numIterations]} \Loss{S_V'}{h_i}$
			\State \textbf{output} $h_{i^*}$
		\end{algorithmic}

		\noindent\rule{\textwidth}{0.5pt}
		
		\vspace*{0.5em}
		
		\textsc{Prover}$(S_P)$:
		\begin{algorithmic}
			\vspace*{0.2em}
			%			\State $\mP \gets \todo{}$
			%			\vspace*{0.5em}
			%			\State $\bigl((x_1,y_2),\dots,(x_{\mP},y_{\mP})\bigr) \gets \cD^{\mP}$
			%			\vspace*{0.5em}
			%			\Comment{Take i.i.d.\ samples from $\cD$}
			%			\Loop \textbf{ forever}:
			%			\vspace*{0.2em}
			%			\State $q^1,\dots,q^k \gets$ \textbf{receive} queries from verifier
			%			%\Comment {Query is a function $\Omega \rightarrow \{0,1\}$}
			%			\vspace*{0.5em}
			%			\State $v^1,\dots,v^k \gets \frac{1}{\mP}\sum_{i \in [\mP]}q^1\left(z_i^{\scriptscriptstyle{P}}\right), \dots, \frac{1}{\mP}\sum_{i \in [\mP]}q^k\left(z_i^{\scriptscriptstyle{P}}\right)$
			%			\vspace*{0.5em}
			%			\State \textbf{send} $v^1,\dots,v^k$ to verifier
			%			\EndLoop 
			\Loop \textbf{ forever}:
			\vspace*{0.2em}
			\State $q \gets$ \textbf{receive} query from verifier
			%\Comment {Query is a function $\Omega \rightarrow \{0,1\}$}
			\vspace*{0.5em}
			\State $v \gets \frac{1}{\mP}\sum_{i \in [\mP]}q\left(\samplepoint{i}{P}\right)$
			\vspace*{0.5em}
			\State \textbf{send} $v$ to verifier
			\EndLoop 
		\end{algorithmic}
		
	\end{ShadedAlgorithmBox}
	
	\caption{A PAC verification protocol for statistical query algorithms.}
	\label{protocol:sq-verification}
\end{protocol}

\begin{protocol}[H]
	\begin{ShadedAlgorithmBox}
		{\bfseries Assumptions:} As in \cref{protocol:sq-verification}. 
		
		\noindent\rule{\textwidth}{0.5pt}
		
		\vspace*{0.5em}
		
		\textsc{VerifierIteration}$(S_V)$:
		\begin{algorithmic}
			\For $t \gets 1,2,\dots$:
			%\State $S_1,\dots,S_\gets$
			\State \textbf{simulate} $A$ until it sends a batch of queries or produces an output
			\If $A$ sends a batch of queries $\bq_t$:
			\If{$t \geq \batchUpperBound$:}
			\State {\bfseries output} $\reject$ and {\bfseries terminate}
			\EndIf
			%\State $q_t^{\scriptscriptstyle \cup} \gets \lor_{i \in [\numQueries_\timestep]} q_t^i$
			\State $\ba_t \gets \atoms{\sigma(\bq_t)}$
			\State \textbf{send} $\ba_t$ to prover
			\State \textbf{receive} $\tilde{\bp}_t$ from prover
			\State \textsc{IdentityTest}$(S_V, \ba_t, \tilde{\bp}_t,\tau)$
			\State $\tilde{\bv}_t \gets $ evaluations for $\bq_t$ induced by $\tilde{\bp}_t$
			%\State $\hat{v}_t^1,\dots,\hat{v}_t^{\numQueries_\timestep} \gets M_\lambda(\tilde{v}_t^1),\dots,M_\lambda(\tilde{v}_t^{\numQueries_\timestep})$
			\State \textbf{send} $\tilde{\bv}_t$ to $A$
			\ElsIf $A$ produces output $h$:
			\State \textbf{return} $h$
			\EndIf
			\EndFor
		\end{algorithmic}
		
		\vspace*{1em}
		
		\textsc{IdentityTest}$(S_V,\ba_t,\tilde{\bp}_t,\tau)$:
		\begin{algorithmic}
			%			\State $\tilde{\cP} \gets $ distribution over $X_0,\dots,X_a$ induced by $\tilde{v}_1,\dots,\tilde{v}_k,\tilde{v}_{\scriptscriptstyle \cup}$\\
			%			\Comment{Computed using Lemma \todo{}.}
			\For $j \in [\mV]$:
			\State $i_j \gets i\in[|\ba_t|] \text{ such that } a_t^i(\samplepoint{j}{V}) = 1$
			\EndFor
			\State{\textbf{execute} the distribution identity tester of \cref{theorem:tolerant-identity-testing}
				\begin{itemize}
					\item[] with sample $I = (i_1,\dots,i_{\mV})$ to distinguish with 
					\item[]{probability at least $1-\varepsilon\delta/4\batchUpperBound$ between
						\[
						\quad \TV{\tilde{\bp}_t, \bp_t} \leq \frac{\tau}{2\sqrt{|\ba_t|}}, \quad \text{and} \quad \tau \leq \TV{\tilde{\bp}_t, \bp_t}
						\]
					}
					%\item[]{where $\tilde{\cP} = \tilde{p}_1,\dots,\tilde{p}_n$ is the distribution that generated $S'$}
					\item[]{where $\bp_t$ is the distribution that generated $I$}
				\end{itemize}
			}
			\vspace*{0.2em}
			%\If{$\tilde{\cP}_{w,f} \notin \distribution{\{0,1\}^w \times \{0,1\}^2}$ or the tester rejects:}
			\If{identity tester rejects:}
			\State \textbf{output} $\reject$ and \textbf{terminate}
			\EndIf
		\end{algorithmic}
		
	\end{ShadedAlgorithmBox}
	\caption{A subroutine of \cref{protocol:sq-verification}.}
	\label{protocol:sq-verifier-iteration}
\end{protocol}

%\begin{lemma}\label{lemma:sq-verification-lemma-old}
%	In the context of \cref{theorem:sq-verification}, fix a distribution $\cD \in \bbD$ and assume $A$ has access to an oracle $\cO_\cD$ as defined in \cref{algorithm:dp-oracle}. Consider an execution of \textsc{VerifierIteration} (\cref{protocol:sq-verifier-iteration}) in which the verifier does not reject, and the query evaluations $\tilde{\bp}_t$ provided by the prover satisfy
%	\begin{align}\label{eq:assumption-prover-honest-enough-for-learning-old}
%	\forall t \in [\finalBatch]: ~ \left\|\tilde{\bp}_t  - \bp_t\right\|_1 \leq \tau,
%	\end{align}
%	where $\bp_t$ is the vector of correct evaluations $p_t^i = \EEE{Z \sim \cD}{a_t^i(Z)}$. Then the output $\hVI \in \cH$ returned by \textsc{VerifierIteration} satisfies
%	\begin{equation}\label{eq:hVI-low-loss-old}
%	\PP{\Loss{\cD}{\hVI} \leq \Loss{\cD}{A} + \frac{\varepsilon}{2}} \geq \frac{\varepsilon}{8}.
%	\end{equation}
%\end{lemma}

\section{Concentration of Measure}

\begin{theorem}[\citealp{hoeffding1963probability}]
	Let $a,b,\mu \in \bbR$ and $m \in \bbN$. Let $Z_1, \dots, Z_m$ be a sequence of i.i.d.\ real-valued random variables and let $Z=\frac{1}{m} \sum_{i=1}^m Z_i$. Assume that $\EE{Z}=\mu$, and for every $i \in [m]$, $\PP{a \leq Z_i \leq b}=1$. Then, for any $\varepsilon>0$,
	\[
		\PP{\left|Z-\mu\right|>\varepsilon} \leq 2 \exp\left(\frac{-2 m \varepsilon^2}{(b-a)^2}\right).
	\]
\end{theorem}

\end{document}